\documentclass{article}
\usepackage[top=1in, bottom=1in, left=1in, right=1in]{geometry}
\usepackage[utf8]{inputenc}
\usepackage{graphicx}
\usepackage{pgfplots}
\fontsize{16pt}{14pt}\selectfont
\title{\textbf{Memory Efficient Adaptive Attention for Multiple Domain Learning}}

\author{Himanshu Pradeep Aswani, Abhiraj Sunil Kanse, Shubhang Bhatnagar, Amit Sethi \\ Indian Institute of Technology, Bombay}

\usepackage[
backend=biber,
style=alphabetic,
sorting=ynt
]{biblatex}

\begin{document}
\maketitle

\fontsize{12pt}{14pt}\selectfont

\begin{center}
\section*{Abstract}
Training CNNs from scratch on new domains typically demands large numbers of labeled images and computations, which is not suitable for low-power hardware. One way to reduce these requirements is to modularize the CNN architecture and freeze the weights of a heavier module -- i.e., the lower layers -- after pre-training. Recent studies have proposed alternative modular architectures and schemes that lead to a reduction in the number of trainable parameters needed to match the accuracy of fully fine-tuned CNNs on new domains. Our work suggests that a further reduction in the number of trainable parameters by an order of magnitude is possible. Furthermore, we propose that new modularization techniques for multi-domain learning should also be compared on other realistic metrics, such as the number of interconnections needed between the fixed and trainable modules, the number of training samples needed, the order of computations required and the robustness to partial mislabeling of the training data. On all of these criteria, the proposed architecture demonstrates advantages over or matches the current state-of-the-art. 
\end{center}

\section{Introduction}
%Paragraph 1 discusses the need for efficient transfer learning/sharing of learnt features.
Over the past decade, deep convolutional neural networks (CNNs) have achieved remarkable results on a wide variety of image analysis tasks, such as classification, object detection, semantic segmentation, and enhancement. However, depending on the domain, CNNs require more than $10^{4}$ labeled training images and upwards of $10^{9}$ floating point operations (FLOPs) to train from scratch (random weight initialization). Clearly, training CNNs from scratch is unsuitable for resource-constrained settings, such as edge devices, robotics, and space exploration. 

To reduce the number of training samples and computations, one can start with a network that is pre-trained on a large dataset, such as ImageNet~\cite{IN}, and replace the last layer with a block of randomly initialized custom layers. After this, either the whole network or only the final few layers are fine-tuned ~\cite{HT}. Modular fine-tuning of only a subset of layers and weights instead of the entire network has the additional advantage of being suitable for implementation on hybrid hardware for low-power applications. For instance, the larger module of the neural architecture with the frozen weights can be implemented on an application specific integrated circuit (ASIC) or field programmable gate array (FPGA), while only the smaller trainable module needs to be implemented on a graphics or central processing unit (GPU or CPU). FPGA and ASIC implementation schemes for CNNs are out of scope for this study, but a comprehensive survey is given in~\cite{capra2020updated}. Our focus is on further reducing the number of parameters needed in trainable modules while learning a new domain.

\begin{figure*}
    \centering
    \includegraphics[trim=1.5cm 7.5cm 1.5cm 1.5cm, clip, width=.6 \textwidth]{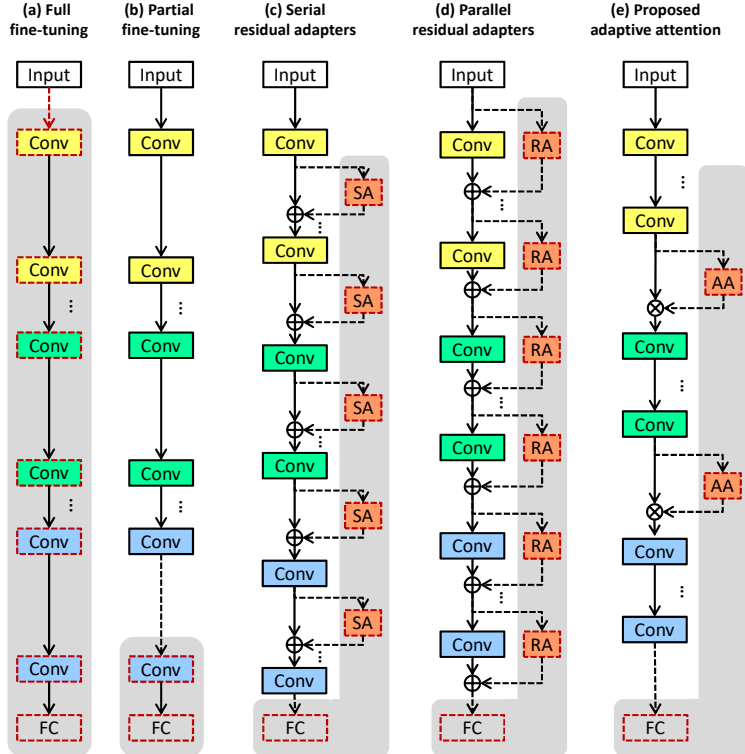}
    \caption{A simplified view of different modular schemes to adapt to new domains, including (a) fine-tuning all layers, (b) fine-tuning only the last few layers, (c) serial residual adapters~\cite{Rebuffi1}, (d) parallel residual adapters~\cite{Rebuffi2}, and (e) adaptive attention (proposed). Blocks with dashed outlines and grey backgrounds are trainable, while others are frozen. Point-wise addition and multiplication (attention) are shown with $+$ and $\times$ respectively. Connections that straddle frozen and trainable modules are shown as dashed arrows. Convolutional layer blocks that differ in feature map dimensions are shown in different colors. The numbers of fully connected (FC) and convolutional (conv) blocks and their form can vary depending on the architecture. Residual connections and other types of layers may also exist (not shown).}
    \label{fig:Types1}
\end{figure*}

\begin{figure*}
    \centering
    \includegraphics[trim=0cm 0cm 0cm 0cm, clip, width=.8\textwidth]{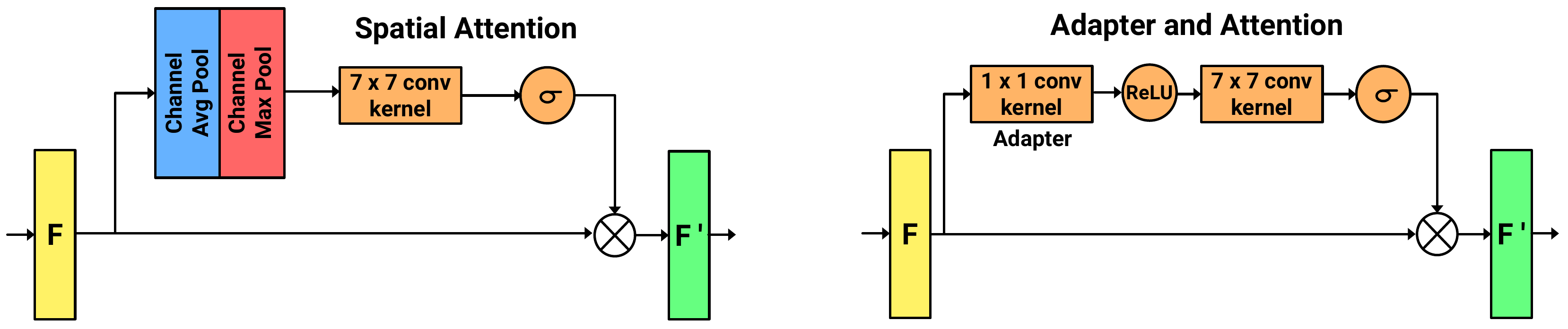}
    \caption{Spatial attention~\cite{CBAM}, and adaptive attention (proposed) modules. The latter is a sequential composition of adapter and attention modules.}
    \label{fig:MD}
\end{figure*}

To reduce the number of trainable parameters, recent studies have departed from the schema of fine-tuning the last few layers by proposing alternative CNN layers~\cite{DepthwiseConv}, training schemes~\cite{MPNAS}, and adapter modules~\cite{Rebuffi1,Rebuffi2}. For benchmarking, these techniques were tested for multi-domain image recognition on the \emph{visual decathlon challenge} (VDC)~\cite{Rebuffi1}.  We depict a further reduction, by \textit{at least an order of magnitude}, in the number of trainable parameters is possible by using the proposed `adaptive attention' modules, while matching the alternatives on the VDC. Inspired by the use of attention to improve the generalization of widely-used CNN architectures~\cite{mnih2014recurrent,SE,wang2017residual,CBAM}, the proposed trainable modules implement spatial attention preceded by a novel feature reduction adapter unit. See Figures~\ref{fig:Types1} and~\ref{fig:MD}. The proposed modules can be seamlessly inserted in a variety of baseline architectures, making them widely applicable.

Our contributions can be summarized as follows:
%\vspace{-0.3cm}
\begin{enumerate}
    \item Demonstrate that domain-specific modules based on attention achieve appreciable performance while reducing the memory requirements by at least 10 times
    \vspace{-10pt}
    \item Integrate the complementary benefits residual adapters and spatial attention generate, termed adaptive attention, to further improve performance on the VDC
    \vspace{-10pt}
    \item Perform a comprehensive comparison of multi-domain learning techniques on a variety of metrics
\end{enumerate}

\section{Related Work}

Previous works that have contributed to or are related to the ideas presented here include those on transfer learning with freezing of lower layers, multi-task and multi-domain learning with weight sharing, discoveries of modularity in neural networks, neural architecture search, incremental learning, and attention in CNNs.

\subsection{Transfer learning and weight freezing}
Some pioneering works on transfer learning that demonstrate pre-training on ImageNet~\cite{IN}, followed by freezing of the weights in lower layers, leads to surprisingly good performance on new domains and tasks are ~\cite{HT,sharif2014cnn}. This method has now become a popular first choice among CNN training schemes for new applications. Besides reducing the training data size and computations, weight freezing has the added advantage of allowing hybrid implementations of CNNs in resource constrained settings. Frozen portions of the CNN architecture may be implemented on a low-power and low-space ASIC. In contrast, the trainable portion of the CNN can be implemented on a GPU ~\cite{capra2020updated}.

\subsection{Multiple task and multiple domain learning with weight sharing}

Multiple-task learning (MTL) has been broadly used to refer to a combination of two or more of the following: segmentation, detection, estimation or resolution, usually on the same training data. Weight sharing is quite common for such tasks, where the set of shared lower layers is called an \emph{encoder, feature extractor, embedding}, or \emph{trunk}, while the task-specific upper layers are called \emph{decoders} or \emph{heads}~\cite{gong2019comparison}. Each head has its own loss function, whose gradients are used to tune not only the weights of that particular head, but also train the shared weights closer to the input. Although there are some usable ideas in MTL, it is not our target problem.

Our goal specifically is multiple-domain learning (MDL). In MDL, the task can remain the same but the image domain, along with its background, changes. For instance, the task can be classification, but the domain can be types of, either Flowers or Aircrafts. The idea of joint training of the trunk in the multiple task setting has been extended to multiple domain learning. For example, the training data between the different domains is usually disjoint, but sharing the encoder trunk makes them robust to training on small datasets with $10^3$ to $10^5$ images~\cite{Rebuffi1,Rebuffi2,MPNAS,Liu_2019_CVPR}. 
New architectures and notions of trainable modules that depart from a layer-wise view include series~\cite{Rebuffi1} and parallel adapters~\cite{Rebuffi2}. Starting with a pre-trained ResNet-style architecture~\cite{he2016deep} that is pre-trained on ImageNet~\cite{IN}, these lightweight trainable adapters are added to each frozen convolutional layer (also see Figure~\ref{fig:Types1}). In another direction, soft sharing of convolution layers -- when they are represented as separable sequential kernels -- has also been explored \cite{DepthwiseConv}. Another insight proposed in \cite{NADAEC} is that the complexity of different domains to be learned might be different. Therefore, branching out from the baseline architecture at strategic points, where specially designed exit modules determine the image class, might be beneficial. We present an even lighter module that achieves competing performance to~\cite{Rebuffi1,Rebuffi2}.

The notion of adding a separate block for each task and designing an explicit neuron augmentation scheme has also been proposed~\cite{KT}. The main principle here is that one may add a parallel layer of neurons to an existing layer and create connections from trained nodes to the newly incorporated ones. In \cite{superposition}, an interesting algorithm of simultaneous storage of multiple models that utilize the same architecture has been put forth. It relies on the construction of a set of basis matrices that spans the weight matrix space for each pair of adjacent layers. Each learned weight matrix can simply be stored as a one-dimensional vector of coefficients corresponding to the basis matrices. This addresses the issue of storage constraints to a large degree. The notion of explicitly designing multiple module combinations for the task of image classification has been formulated in \cite{taskdecomposition}.

The task of multi-domain learning is not always approached from the point of view of freezing a large section of the weights. We consider comparisons with techniques where the shared weights are jointly trained for different tasks or domains, such as~\cite{Liu_2019_CVPR}, to be out of scope. Our explicit goal is to propose architectures and training schemes where the shared weights are \emph{pre-trained and frozen} without being influenced by images from the future domains on which they will be deployed. This restriction is necessary to test whether additional lightweight trainable modules are sufficient to give high task- or domain-level accuracy. This can potentially allow heavier frozen modules to be implemented on optimized hardwares for resource-constrained settings, and shipped frozen in ``factory settings'', so to speak.

\begin{figure*}
    \centering
   \includegraphics[scale=0.7]{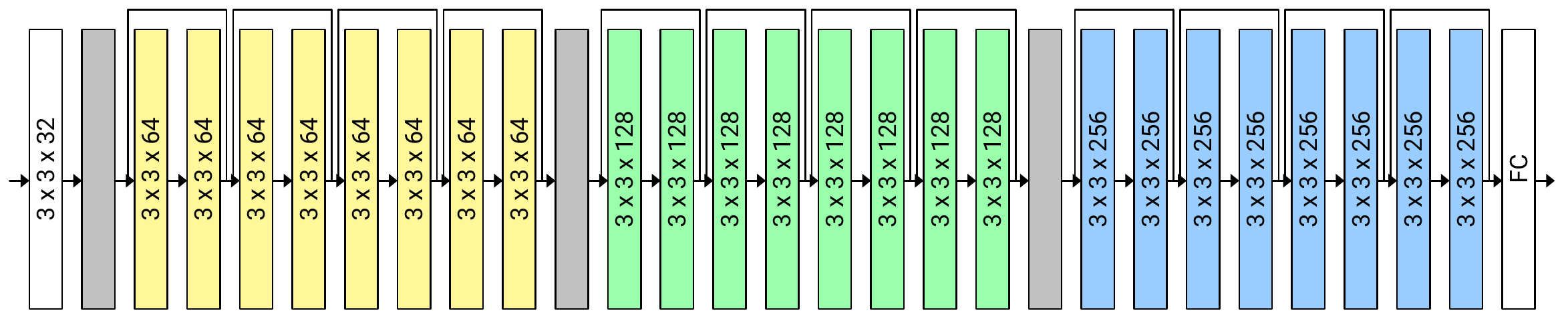}
   \caption{The ResNet26 baseline architecture, as proposed in \cite{Rebuffi1}. Grey blocks denote the positioning of our modules within the larger network. The FC layer is replaced for every new task and is not included in the domain-specific parameter count. Best viewed in color.}
\label{fig:BaseNetwork}
\end{figure*}

\subsection{Neural architecture search}
Neural architecture search refers to exploring the space of possible neural architectures that allow for the best possible performance on a particular task. Principal approaches in this area involve optimizing on the number and shape of each layer to be included in the final network design. Traditionally, these techniques are computationally expensive due to their requirements of multiple iterations of training over the same dataset to arrive at the optimal structure. Modern approaches, such as~\cite{FP}, aim to determine the utility of each learned convolutional layer in a pre-trained model by constructing binary masks for each new task. An exploration of three possible scenarios: reuse, adapt, or have a totally separate block of neurons to be trained from scratch for every layer of a neural network, whenever a new domain is to be learned, was also carried out~\cite{thesis}. An algorithm for estimating the exact structure of the network given a budget in terms of number of neuron units allowed is designed in \cite{Lee_2020_CVPR}. Neural architecture search, which attempts to discover an optimal route of information flow within a super graph for a particular domain, has also been proposed and tested on Visual Decathlon Challenge~\cite{MPNAS}, to which we compare our results.

\subsection{Incremental learning}
Popular incremental learning techniques require one to learn the sub-tasks of a particular problem sequentially. There are approaches, such as \cite{Kirkpatrick3521} and \cite{LWF}, that attempt to tailor the neural network to perform well on the earlier as well as latter tasks by updating the weights to serve all tasks, by utilizing only new task data. These are not directly compared with our work as they do not advocate a frozen backbone as well as outline an explicit domain-specific modular scheme along the length of the neural network. Others mandate maintaining a repository of previous training material to ensure that the network does not forget the previously learned sub-tasks. To improve these type of upon current incremental learning algorithms, \cite{Castro_2018_ECCV} proposes reducing the memory footprint by constraining the size of repository one can access. It was suggested to further build upon this idea by defining two metrics one should look out for, namely, forgetting and intransigence \cite{Chaudhry_2018_ECCV}. The main objective here is to maintain these parameters at desirable values when learning new tasks. The requirement to maintain samples from a previous domain or sub-tasks make these techniques out-of-scope for a direct comparison with our work, as we would prefer the modular CNNs to be ready for use with limited tuning on a single domain.

\section{Proposed Adaptive Attention Module}
% Paragraph 0 presents the design of Adapters. 
The inspiration for the proposed adaptive attention design comes from complementing domain-wise adapters~\cite{Rebuffi1,Rebuffi2} with the channel and spatial attention operations that constitute the convolutional block attention module a.k.a CBAM(~\cite{CBAM}), with additional modifications.

The concept of deploying domain-specific adapters to improve multiple domain learning was first proposed by \cite{Rebuffi1}. The possibility of employing adapters in parallel with every convolution layer in the neural network has been further explored in \cite{Rebuffi2}, whose results improve upon those of~\cite{Rebuffi1} without increasing the number of trainable parameters. In~\cite{Rebuffi2}, every convolutional layer is paired with a custom adapter and is represented mathematically as follows:
\begin{equation}\label{1}
F' = F \star K + F \star \alpha,
\end{equation}
where $F  \in   R^{C \times H \times W}$ is the input, $K  \in   R^{C \times k \times k}$ is the core convolution kernel, $\alpha  \in  R^{C \times 1 \times 1}$ is a convolution block that is referred to as an adapter, $\star$ represents the convolution operation, and $F'$ represents the output of that layer in the neural network. Note that the $\star$ represents convolution operations with appropriate stride and padding values to ensure additive compatibility.
%Paragraph 1 describes the CBAM structure.

Channel and spatial attention blocks have been used for reweighing the input channels and spatial groups of input pixel positions as per their importance~\cite{SE,CBAM}. Attention weights can improve task performance with the addition of a relatively small (roughly $1\%$) fraction of parameters. Given an input tensor $F  \in  R^{C \times H \times W}$, the channel attention block computes a 1-D channel attention map $M_{c}  \in  R^{C \times 1 \times 1}$ and the spatial attention block computes a 2D spatial attention map $M_{s}  \in  R^{1 \times H \times W}$. The complete sequence of operations for computing attention, as proposed in \cite{CBAM}, is as follows:
\begin{equation}\label{2}
F' = M_{c}(F) \otimes F,
\end{equation}
\vspace{-20pt}
\begin{equation}\label{3}
F'' = M_{s}(F') \otimes F',
\end{equation}
where $\otimes$ represents the element-wise multiplication. 

As one compares the forms of Equations~\ref{1} and~\ref{3}, one notices that the convolution operation with $\alpha$ in Equation~\ref{1} basically represents multiple linear combinations of the input channels without any regards to variation along the spatial dimensions, unlike Equation~\ref{3}. On the other hand, Equation~\ref{2} is \textit{fundamentally a special case} of the adapter($\alpha$) in Equation~\ref{1}, as channel attention represents only one linear combination. We therefore conclude that Equation~\ref{1} and Equation~\ref{3} capture different operations which could potentially complement each other. The adaptive attention module design is based on this insight.

\subsection{Intermediate module design - Spatial Attention}
To ascertain the potential in deploying spatial attention as a domain-specific module, we first simply augment the base network with the spatial attention module ~\cite{CBAM}. The spatial attention module can be mathematically expressed as follows:
\begin{equation}\label{4}
    M_{s}(F) = \sigma(f \star ([AvgPool(F),MaxPool(F)])),
\end{equation}
where $\sigma$ denotes the sigmoid operation, pooling is across the channel dimensions so that AvgPool(F) $\in R^{1 \times H \times W}$, MaxPool(F) $\in R^{1 \times H \times W}$, convolution kernel $f \in  R^{2 \times k \times k}$ ($k=7$, as per ~\cite{CBAM}) and [] represents the concatenation operation.

Although, such an attention module has been used to improve the accuracy of a CNN~\cite{CBAM}, \textit{these modules have not been used previously as the only trainable module to test lightweight adaptation for multiple domain learning}.

\subsection{Complete module design - Adaptive Attention}\label{5}
Now, we aim to combine the mathematical advantages of both the adapter and the attention designs along the spatial dimensions to construct the proposed adaptive attention module, which can be formally described as follows:

\begin{equation}
    F' = F \otimes \sigma(K \star ReLU(F \star \alpha)),
\end{equation}
where $\sigma$ denotes the sigmoid operation, ReLU stands for the rectified linear unit activation function, $F  \in  R^{C \times H \times W}$ is the input, $\alpha  \in  R^{C \times 1 \times 1}$ is the adapter, and $K\in  R^{C \times k \times k} (k=7),$ is the spatially attentive convolutional kernel, playing the role of f in Equation \ref{4}.

\section{Experiments and Results}

We evaluate our proposed module architecture as well as fine-tuning, only final convolution layer trainable and residual adapters approach on the newly introduced Visual Decathlon Challenge. Official submissions to the online test server receive per-task accuracies as well as a cumulative score value out of 10,000. The objective is to maximize the total score while simultaneously ensuring a smaller memory footprint. The score is formulated as a weighted combination of test accuracies achieved, with the weights capturing the level of difficulty in learning the dataset. Further details of the challenge can be found at www.robots.ox.ac.uk/~t vgg/decathlon/.

\begin{table}
\large
\label{tab:ResultsResNet26}
%\vspace{0.1cm}
%\begin{tabular}{|l|c|c c c c c c c c c c |c|c|}
\begin{tabular}{|p{1.4 cm}|p{1cm} p{0.8cm} p{0.7cm} p{0.7cm} p{0.7
cm} p{0.7cm} p{0.7cm} p{0.7cm} p{0.7cm} p{0.8cm} p{0.7cm} p{0.7cm} p{0.9cm}|}
\hline
Method & $\% Tun$ & ImNet & Airc. & C100 & DPed & DTD & GTSR & Flwr & OGlt & SVHN & UCF & Mean & Score\\
%\hline
$\#$images &  & 1.3m & 7k & 50k & 30k & 4k & 40k & 2k & 26k & 70k & 9k & & \\
\hline
FT & 100 & 60.3 & 43.7 & 80.7 & 96.3 & 51.2 & 99.1 & 71.7 & 86.0 & 96.2 & 48.4 & 73.4 & 2642\\

Conv & 10.28 & 60.3 & 36.6 & 77.9 & 97.1 & 52.1 & 97.4 & 68.7 & 82.7 & 94.2 & 40.6 & 70.8 & 1808\\

P.R.A.* & 10.11 & 60.3 & 20.7 & 78.2 & 97.6 & 39.4 & 98.6 & 41.1 & 87.3 & 91.1 & 33.1 & 64.7 & 2050\\

Org. P.R.A. & 10.11 & 60.3 & 64.2 & 81.9 & 94.7 & 58.5 & 99.3 & 84.7 & 89.2 & 96.5 & 50.9 & 78.1 & 3412\\

\hline
S.Atn. & 0.14 & 60.3 & 46.8 & 76.8 & 97.4 & 48.5 & 97.5 & 71.3 & 82.4 & 92.9 & 44.6 & 71.9 & 1899\\

Ad.Atn.& 0.15 & 60.3 & 43.4 & 77.7 & 96.9 & 50.1 & 99.7 & 70.3 & 83.6 & 94.8 & 44.2 & 72.1 & 2565\\
\hline
\end{tabular}
\caption{Top-1 Accuracy (averaged over three random seeds) of different MDL approaches obtained on the Visual Decathlon Challenge online server using pre-trained ResNet26~\cite{he2016deep,Rebuffi1}. FT: finetune Conv: Final convolution layer trainable P.R.A.: parallel residual adapters S.Atn.: Spatial Attention Ad.Atn.: Adaptive Attention Org.: Original  *: our implementation. Differences may be due to hyper-parameters}
\end{table}

\begin{table}
\large
\label{tab:ResultsRMobileV3}
%\vspace{0.1cm}
%\begin{tabular}{|l|c|c c c c c c c c c c |c|}
\begin{tabular}{|p{1.4 cm}|p{1cm} p{0.8cm} p{0.7cm} p{0.7cm} p{0.7
cm} p{0.7cm} p{0.7cm} p{0.7cm} p{0.7cm} p{0.8cm} p{0.7cm} p{0.7cm} p{0.9cm}|}
\hline
Method & $\% Tun$ & ImNet & Airc. & C100 & DPed & DTD & GTSR & Flwr & OGlt & SVHN & UCF & Mean & Score\\
%\hline
$\#$images  &  & 1.3m & 7k & 50k & 30k & 4k & 40k & 2k & 26k & 70k & 9k & & \\
\hline
MPNAS & 3.00 & 57.0 & 29.7 & 81.2 & 96.0 & 36.6 & 99.8 & 79.2 & 74.2 & 92.3 & 71.8 & 71.8 & 2677\\
\hline
Ad.Atn. & 0.85 & 74.3 & 29.4 & 70.0 & 96.2 & 46.5 & 97.8 & 61.5 & 82.5 & 93.7 & 69.8 & 72.2 & 2075\\
\hline
\end{tabular}
\caption{Top-1 Accuracy (averaged over three random seeds) of different MDL approaches obtained on the Visual Decathlon Challenge online server}
\end{table}

\subsection{Training setup}
All experiments run herein have the weights of the baseline architectures, which correspond to the core convolution layers being frozen after training on ImageNet~\cite{IN}. The images of every dataset have been resized to $72 \times 72$ to ensure a uniform starting point for all MDL schemes. All experiments have been executed for a maximum of 75 epochs. Typical weight decay values chosen were \{$10^{-2}$,$10^{-3}$,$10^{-4}$\}. Learning rates were chosen depending on the number of trainable parameters from \{$5\times 10^{-1}$,$10^{-1}$,$5\times 10^{-2},10^{-2}$\} for the SGD optimizer and \{$10^{-2}$,$10^{-3}$,$10^{-4}$\} for the Adam optimizer. We have chosen two base architectures, ResNet26 and MobileNetV3 as our reference and compare our results to only those techniques that utilize the same reference.

\subsection{Comparison with Parallel Residual Adapters using ResNet26}
Parallel~\cite{Rebuffi2} residual adapters are implemented on a ResNet26 baseline architecture. This customized architecture differs significantly from the standard ResNet series of neural networks~\cite{he2016deep} in terms of the number of convolutional layer blocks. Colored blocks as well as the starting and ending white blocks, in Figure~\ref{fig:BaseNetwork}, depict the ResNet26 architecture. Evidently, it consists of three broad stages of convolution layers. \textit{We incorporate our module before each stage, highlighted by the grey colored blocks} in Fig \ref{fig:BaseNetwork}, and learn different weights for these modules for every new domain.

Table~\ref{tab:ResultsResNet26} presents the results of this experiments. Our approach can match or slightly exceed the accuracy of parallel residual adapters~\cite{Rebuffi2} with approximately \textit{two orders of magnitude fewer  parameters}. For certain datasets, such as Aircrafts (\textbf{Airc.}), DTD, Flowers(\textbf{Flwr}) and Human Actions(\textbf{UCF101}) the adaptive attention module achieves a much higher accuracy than the residual adapters approach. Apart from the Omniglot(\textbf{OGlt}) dataset, we achieve comparable results on remaining datasets in the VDC.

\begin{table}
\large
\label{tab:ParamCalc}
\begin{tabular}{|p{2.4cm}|p{1cm} p{1cm} p{1.4cm} p{1.4cm} p{1.4cm} p{1.4cm} p{1.3cm} p{1.4cm}| }
\hline
Method & Giga & $\Delta$  & Total & Train. & Total & Train. & Train.  & Conn.s\\
 & \multicolumn{2}{c}{MACs}  & & & w/o FC & w/o FC & \% & \\
\hline
FineTune & 10.48 & 0 & 5,827 k & 5,827 k & 5,815 k & 5,815 k & 100.00 & 16 k \\
Final Conv.& 10.48 & 0 &5,827 k & 610 k & 5,815 k & 597 k & 10.28 & 21 k\\

Par.Res.Adpt. & 11.64 & 1.16 & 6,472 k & 665 k & 6,460 k & 653 k & 10.11 & 2,249 k \\

Spa. Attn. & 10.49 & 0.01 &5,827 k & 20 k & 5,816 k & 8 k & 0.14 & 297 k\\

Adpt. Attn. & 10.49 & 0.01&5,827 k & 21 k & 5,816 k & 9 k & 0.15 & 297 k\\
\hline
\end{tabular}
\caption{An example of tallying up the number of total and trainable parameters as well as MACs (Multiply-Accumulates) for the Describable Textures Dataset(DTD).}
\end{table}

\subsection{Comparison with MPNAS using MobileNetV3}
%Paragraph 1 discusses the approach involved in MPNAS. 
The general approach in neural architecture search (NAS) typically involves the construction of a super graph consisting of multiple nodes, each representing a convolution layer. Multiple Path Neural Architecture Search (MPNAS)~\cite{MPNAS} is a reinforcement learning-based path routing mechanism on the MobileNetV3-like search space. We, therefore, use the MobileNetV3 Large neural network~\cite{Howard_2019_ICCV} as our baseline architecture to maintain a similar number of learnt parameters. A careful analysis of the backbone network reveals four broad stages of layers, the characteristic feature being the kernel size. \textit{We position one module after each of the four stages}, freeze the base network and keep the batch normalization layers trainable.

%For the comparison with MPNAS~\cite{MPNAS}, we state their results directly, and search over the same 18 hyper-parameter combinations as mentioned for the comparison with residual adapters~\cite{Rebuffi2} on the ResNet26 baseline, as described in the previous sub-section, for our adaptive attention module design. 

Table~\ref{tab:ResultsRMobileV3} presents the accuracy our methodology obtains when incorporated in the MobileNetV3 architecture~\cite{Howard_2019_ICCV}, on the VDC ~\cite{Rebuffi1}. When compared to the approach proposed in MPNAS~\cite{MPNAS}, our method requires a fraction of the total number of  parameters. Additionally, our approach is much simpler in terms of execution complexity as we do not depend upon reinforcement learning based algorithms to design our architectures. Furthermore, we are not required to discover paths for every new domain in the super graph, thus avoiding computationally-heavy re-training. On a few particular datasets, such as DTD and Omniglot(\textbf{OGlt}), we outperform the MPNAS approach by a large margin. On other datasets, such as CIFAR100(\textbf{C100}) and Flowers(\textbf{Flwr}), their approach leads to slightly better results than ours\footnote[1]{Validation accuracy for UCF101 dataset in Table \ref{tab:ResultsRMobileV3} due to discrepancies between validation and test sets, as also reported by others.}.

\subsection{Parameter count and MACs}
The storage of weights and biases learned during training represent a major cost of implementing machine learning models on the edge. The major components of a deep neural network typically comprise convolutional kernels for feature extraction, batch normalization parameters for stable training, and fully connected (FC or dense) layers at the top of the neural network with task specific designs. Therefore, the number of parameters related to the FC layers can range from 1k (for the DPed dataset) to 256k (for the ImageNet challenge). As the core backbone typically has 5.8 million parameters in case of ResNet26 and 3 million in case of MobileNetV3, it seems unreasonable to include FC parameter counts while computing fraction of incremental storage required. We therefore compute the parameter counts as follows. All parameters excluding FC weights and biases shall be our base count. Module architectures should be judged as a fraction of this. The column $\% Tunable$ in Tables~\ref{tab:ResultsResNet26} and~\ref{tab:ResultsRMobileV3} refers to the additional fraction of parameters required with 100\% representing the base count defined earlier. This is consistent with broader definitions in literature and differs only in the aspect of stating percentages per dataset. As an example, we demonstrate calculations involved for the DTD dataset in Table \ref{tab:ParamCalc}. Note that the entries in columns titled w/o FC as well as Train. $\%$ are constant, irrespective of the dataset.

Another metric that serves as a primary check on hardware utilization is the number of MACs (multiply-accumulate) of the complete model. One may observe from Table \ref{tab:ParamCalc} that the increase in MACs is 0.01G for the adaptive attention technique whereas it is 1.16G for the parallel adapters technique, a 100-fold decrease in the additional MACs incurred. Thus, our technique is not only storage efficient but also improves upon the MACs requirement.

\subsection{Number of interconnections}\label{sec:NOI}
Another attribute on which different MDL approaches should be compared is the ease of implementation on a hybrid hardware in terms of the number of interconnections required between the frozen and the trainable modules. A large number of connections in and out of the frozen trunk (e.g., when implemented as an ASIC) complicates the hardware implementation further, apart from the complexity of the trainable module (e.g., the one implemented on a GPU). We present, in Table~\ref{tab:ParamCalc}, values for the number of tap-ins and tap-outs required for each approach under the Connections column. For the parallel residual adapters and our methods, some feature maps are sent back to the fixed module as well. The direction of information flow on these connections may reverse during back propagation. Refer to Figure~\ref{fig:Types1} for a visualization of these interconnections (dashed arrows). We only need 0.30 million interconnections as opposed to 2.25 million because the proposed trainable adaptive attention modules are sparingly distributed in the ResNet26 architecture, as opposed to using it for every convolutional layer (as is done for residual adapters~\cite{Rebuffi2}).
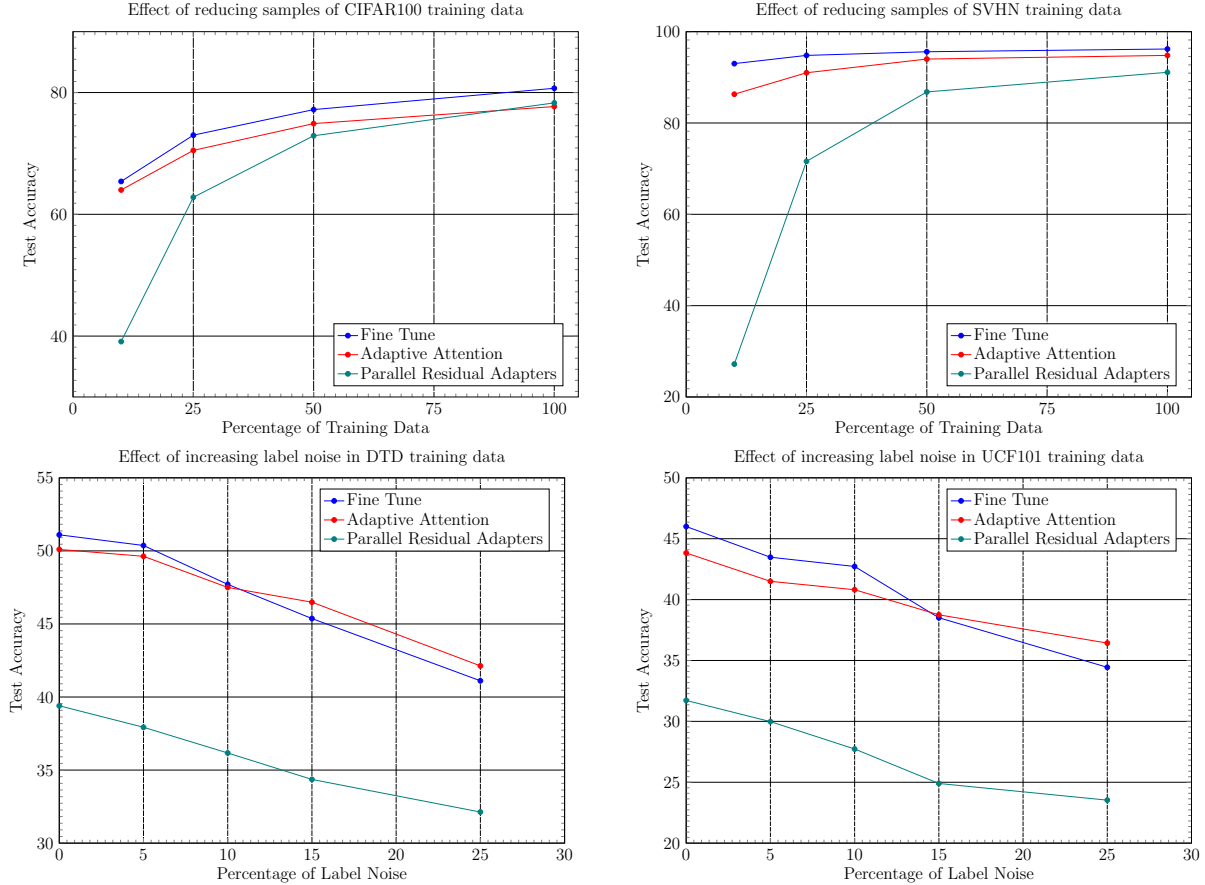
\begin{figure}
\begin{center}
\pgfplotstableread{cifar100.dat}{\table}
\begin{tikzpicture}[scale=0.45]
\Large
\begin{axis}[
    title={Effect of reducing samples of CIFAR100 training data},
    xlabel={Percentage of Training Data},
    ylabel={Test Accuracy},
    xmin = 0, xmax = 105,
    ymin = 30, ymax = 90,
    xtick distance = 25,
    ytick distance = 20,
    grid = both,
    minor tick num = 10,
    major grid style = {black},
    minor grid style = {white},
    width = \textwidth,
    height = 0.75\textwidth,
    legend cell align = {left},
    legend pos = south east
]

\addplot[blue, mark=*] table [x = {x}, y = {y1}] {\table};
\addplot[red, mark=*] table [x ={x}, y = {y2}] {\table};
\addplot[teal, mark=*] table [x = {x}, y = {y3}] {\table};
\legend{
    Fine Tune, 
    Adaptive Attention,
    Parallel Residual Adapters
}
\end{axis}
\end{tikzpicture}
\hspace{10pt}
\pgfplotstableread{svhn.dat}{\table}
\begin{tikzpicture}[scale=0.45]
\Large
\begin{axis}[
    title={Effect of reducing samples of SVHN training data},
    xlabel={Percentage of Training Data},
    ylabel={Test Accuracy},
    xmin = 0, xmax = 105,
    ymin = 20, ymax = 100,
    xtick distance = 25,
    ytick distance = 20,
    grid = both,
    minor tick num = 10,
    major grid style = {black},
    minor grid style = {white},
    width = \textwidth,
    height = 0.75\textwidth,
    legend cell align = {left},
    legend pos = south east
]

\addplot[blue, mark=*] table [x = {x}, y = {y1}] {\table};
\addplot[red, mark=*] table [x ={x}, y = {y2}] {\table};
\addplot[teal, mark=*] table [x = {x}, y = {y3}] {\table};
\legend{
    Fine Tune, 
    Adaptive Attention,
    Parallel Residual Adapters
}
\end{axis}
\end{tikzpicture}
%\vspace{.5cm}
\pgfplotstableread{dtd.dat}{\table}
\begin{tikzpicture}[scale=0.45]
\Large
\begin{axis}[
    title={Effect of increasing label noise in DTD training data},
    xlabel={Percentage of Label Noise},
    ylabel={Test Accuracy},
    xmin = 0, xmax = 30,
    ymin = 30, ymax = 55,
    xtick distance = 5,
    ytick distance = 5,
    grid = both,
    minor tick num = 10,
    major grid style = {black},
    minor grid style = {white},
    width = \textwidth,
    height = 0.75\textwidth,
    legend cell align = {left},
    legend pos = north east
]

\addplot[blue, mark=*] table [x = {x}, y = {y1}] {\table};
\addplot[red, mark=*] table [x ={x}, y = {y2}] {\table};
\addplot[teal, mark=*] table [x = {x}, y = {y3}] {\table};
\legend{
    Fine Tune, 
    Adaptive Attention,
    Parallel Residual Adapters
}
\end{axis}
\end{tikzpicture}
\hspace{10pt}
\pgfplotstableread{ucf.dat}{\table}
\begin{tikzpicture}[scale=0.45]
\Large
\begin{axis}[
    title={Effect of increasing label noise in UCF101 training data},
    xlabel={Percentage of Label Noise},
    ylabel={Test Accuracy},
    xmin = 0, xmax = 30,
    ymin = 20, ymax = 50,
    xtick distance = 5,
    ytick distance = 5,
    grid = both,
    minor tick num = 10,
    major grid style = {black},
    minor grid style = {white},
    width = \textwidth,
    height = 0.75\textwidth,
    legend cell align = {left},
    legend pos =  north east
]

\addplot[blue, mark=*] table [x = {x}, y = {y1}] {\table};
\addplot[red, mark=*] table [x ={x}, y = {y2}] {\table};
\addplot[teal, mark=*] table [x = {x}, y = {y3}] {\table};
\legend{
    Fine Tune, 
    Adaptive Attention,
    Parallel Residual Adapters
}
\end{axis}
\end{tikzpicture}
\end{center}
\caption{Comparison of various techniques for reduced datasets and increased label noise}
\label{fig:f}
\end{figure}

\subsection{Reducing the number of training samples}\label{sec:FracData}
The principal goal of transfer learning is to get high test accuracy with smaller datasets from the target domain or task. Although the datasets in the VDC are already small, we further tested how different MDL schemes handle the reduction in the number of training samples by training them on 10\%, 25\%, 50\% and 100\% of the data for CIFAR100 and SVHN. As shown in Figure \ref{fig:f}, our technique almost matched the accuracy of fine-tuning for various fractions of the training data, while the accuracy of residual adapters reduced significantly for smaller fractions of the dataset.

\subsection{Label noise}\label{sec:LabelNoise}
Realistically, a certain extent of data mislabeling is to be expected in the training datasets due factors, such as human or NLP-driven labeling errors. To test the robustness of adaptive attention to label noise, we simulated four different levels of mislabeling -- 5\%, 10\%, 15\% and 25\% -- by randomly selecting and mislabeling a proportion of the training samples for the UCF101 and DTD datasets. As shown in Figure \ref{fig:f}, once again, adaptive attention performed on par with fine-tuning with far fewer trainable parameters. On the other hand, residual adapters started with a low accuracy, and degraded even further with label noise. We believe the criss-cross that occurs in both the graphs is due to the fact that having too many parameters, may cause overfitting in the model, in the presence of noise. 

\section{Conclusions and Future Work}
In this paper, we introduced adaptive attention modules for multi-domain learning. The proposed modules are lighter than the state-of-the-art methods in terms of number of trainable parameters, and yet can match them in average accuracy on the Visual Decathlon Challenge~\cite{Rebuffi1}. These modules enable one to efficiently utilize previous knowledge to learn new tasks. 

The advantage of treating only certain components as trainable is that the fixed modules do not need to perform any internal updates. The hardware implementation of the fixed modules can be optimized for forward and backward passes without weight updates, while the lightweight trainable modules can be implemented on a CPU or a GPU. Such hybrid architectures, we believe, hold a lot of promise for low-power applications, including edge devices, robotics, and space exploration~\cite{capra2020updated}.

Additionally, for a careful choice of partially trainable architectures, we advocate that one should not only check the number of trainable parameters and accuracy on new domains, but it may also be important to check robustness to reduction in the size of the training data, and injection of label noise. The proposed modules demonstrate competitive results on these attributes.

The extension of our work to other tasks, such as segmentation or detection, is straightforward. This is due to the ease of integration our technique provides, in comparison to other popular adaptation schemes. Our work suggests that further research in modular neural network architectures, including innovative uses of attention mechanisms, is warranted. Additionally, the reason for the strong performance of adding some attention modules over adding more sequential layers is also not totally clear, and deserves more research. 

\printbibliography

\end{document}